# Ensemble Learning of Colorectal Cancer Survival Rates

Chris Roadknight
School of Computing Science
University of Nottingham Malaysia Campus
Malaysia
Chris.roadknight@nottingham.edu.my

Uwe Aickelin
School of Computing Science
University of Nottingham
UK
Uwe.aickelin@nottingham.ac.uk

John Scholefield
Faculty of Medicine & Health Sciences
The University of Nottingham
UK

Lindy Durrant
Faculty of Medicine & Health Sciences
The University of Nottingham
UK

*Abstract*— **In this paper, we describe a dataset relating to cellular and physical conditions of patients who are operated upon to remove colorectal tumours. This data provides a unique insight into immunological status at the point of tumour removal, tumour classification and post-operative survival. We build on existing research on clustering and machine learning facets of this data to demonstrate a role for an ensemble approach to highlighting patients with clearer prognosis parameters. Results for survival prediction using 3 different approaches are shown for a subset of the data which is most difficult to model. The performance of each model individually is compared with subsets of the data where some agreement is reached for multiple models. Significant improvements in model accuracy on an unseen test set can be achieved for patients where agreement between models is achieved.**

## I. Introduction

Colorectal cancer is the third most commonly diagnosed cancer in the world. Colorectal cancers start in the lining of the bowel and grow into the muscle layers underneath then through the bowel wall [1]. TNM staging involves the Classification of Malignant Tumours

- Tumour (T). Size of the tumor and whether it has invaded nearby tissue
- Nodes (N). The extent to which regional lymph nodes involved
- Metastasis (M). This is the spread of a disease from one organ or part to another non-adjacent organ.

4 TNM stages (I,II,III,IV) are generated by combining these three indicator levels and are allied with increasing severity and decreasing survival rates.

Treatment options include minor/major surgery, chemotherapy, radiotherapy but the correct treatment is heavily dependent on the unique features of the tumour which are summarised by the TNM staging. Choosing the correct treatment at this stage is crucial to both the patient's survival and quality of life. A major goal of this research is to automatically optimize the treatment plan based on the existing data.

The data for this research was gathered by scientists and clinicians at City Hospital, Nottingham. The dataset we use here is made up of over 200 possible attributes for 462 patients. The attributes are generated by recording metrics at the time of tumour removal, these include:

- Physical data (age, sex etc)
- Immunological data (levels of various T Cell subsets)
- Biochemical data (levels of certain proteins)
- Retrospective data (post-operative survival statistics)
- Clinical data (Tumour location, size etc).

In the research into the relationship between immune response and tumour staging there has been some support of the hypothesis that the adaptive immune response influences the behavior of human tumors. In situ analysis of tumor-infiltrating immune cells may therefore be a valuable prognostic tool in the treatment of colorectal cancer [2]. The immune and inflammation responses appear to have a role to play in the responses of patients to cancer [3] but the precise nature of this is still unclear.

This research brings together earlier findings on the nature of the tumour biomarker dataset and builds an ensemble learning solution that offers improved predictive performance for a subset of patients.

## II. Background

The dataset supplied is a biological dataset and as such has a rich complement of pre-processing issues. For instance, over 10% of the values are missing, with some attributes having over 40% missing values and some patients having over 30% missing values. Missing data poses a problem for most modelling techniques. One approach would be to remove every patient or every attribute with any missing data. This would remove a large number of entries, some of which only have a few missing values that are possibly insignificant. Another approach is to average the existing values for each attribute and to insert an average into the missing value space. The appropriate average may be the mean, median or mode depending on the profile of the data, because of this the handling of missing values attributes with a non-linear effect is particularly difficult.

Much of the data takes the form of human analysis of biopsy samples stained for various markers. Rather than raw cell counts or measurements of protein levels we are presented with thresholded values. For instance, CD16 [4] is found on the surface of different types of cells such as natural killer, neutrophils, monocytes and macrophages. The data contains a simple 0 or 1 representing the existence or otherwise of a significant number of these cells rather than a count of the number of cells. This kind of manual inspection and simplification is true for most of the data and any modeling solution must work with this limitation.

By using a combination of correlation coefficients and expert knowledge the data was reduced down to a set of ~50 attributes. This included removing several measurements that were hindsight dependent (ie. chemo or radio treatment) and correlated with TNM stage. (ie. Dukes stage).

The survival of patients with TNM class 2 and 3 tumours, the middle grade categories, is the most difficult to predict, Single attribute relationships exist within the dataset but are not strong. Analysis of single attributes in relation to survival can yield predictions with accuracy of only 59%.

## III. Preprocessing

A range of preprocessing methods were applied to the data beyond interpolating. Some patients had extensive missing values, while some attributes were very poorly sampled. There were also some patients who's operation was too recent or who had died of an unrelated cause before the success of the operation could be evaluated. An initial pass followed the following protocol:

*Removed patients with data for fewer than 50% attributes*

*Removed attributes with data for fewer than 50% patients*

*Removed Patients still alive but for less than 60 months post operation*

*Removed patients who died before the 60 month survival threshold of unrelated reasons*

*Removed TNM related attributes*

*Removed post-operative attributes (eg. Radiotherapy)*

*Removed derivable or correlated attributes*

*Removed compound attributes*

Next an effort was made to convert non-linear, single attribute relationships into linear ones. In an effort to better represent attributes with non-linear effect we also converted attributes with non-linear effects into linear effects. FLIPL is a good example (figure 2), values of 0 and 3 have poorer prognosis for survival than 1 and 2, thereby giving a non-linear relationship between grade and survival. Merging the values for FLIPL of 0 and 3 into a new group '0' and 1 and 2 into another group '1' means that using the mean value for missing values now has a meaningful representation for the resulting modeling approach.

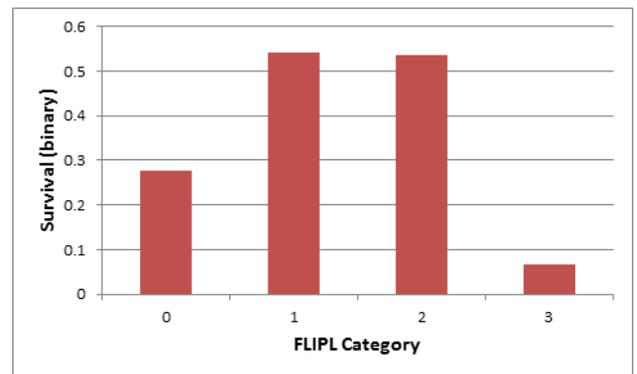

**Figure 2.** Relationship between survival and FLIPL category

A range of supervised [10] and unsupervised [11] method have been applied to this dataset both in predicting TNM classes from immune-histochemical data and predicting survival. The fact that an anti-learning method [10] appears to perform as well as any learning methods for most tasks suggests that there are complex behaviours at work within this dataset, coupled with its high dimensionality, leading to relatively poor prediction results.

## IV. Supervised Ensemble Learning of Survival Rates

The data used in this paper has detailed information about post-operative survival, so we can model how attribute values effect survival rates. For the purposes of this work a patient is deemed to have "survived" if they are still alive 5 years after their tumour removal. We initially looked at how linear transformation and attribute selection affected the ability to predict 5 year survival. Details of the neural network modeling approach are given later in this section but at this stage if we just look at the impact of this reduction in non-linearity. We see a much improved performance with survival prediction levels better than the 65% achieved by TNM alone (figure 3). We can also see that using the top 8 attributes selected by SVM attribute selection [18] provides the best performance.

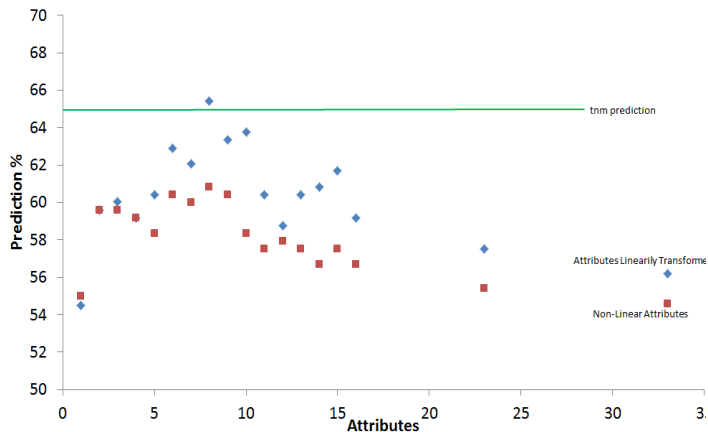
**Figure 3.** Prediction accuracy with an increasing number of attributes with and without non-linearity removal

Several of the attributes presented in the dataset pertain to the survival of the patients after their operation to remove the tumour. The number of months the patient has survived, whether they are still alive or not and how they died (if dead) are all available. Figure 4 shows survival curves for patients up to 6 months. The strong difference between survival rates in TNM stage 1 and 4 patients is apparent (ie. at 30 month the survival rate is approximately 95% and 5% for these 2 groups). The difference between patients with TNM stage 2 and 3 cancers is less apparent. After 30 months deaths from colorectal cancer for TNM stage 1 patients increase quite quickly, in percentage terms steeper than any other TNM class.

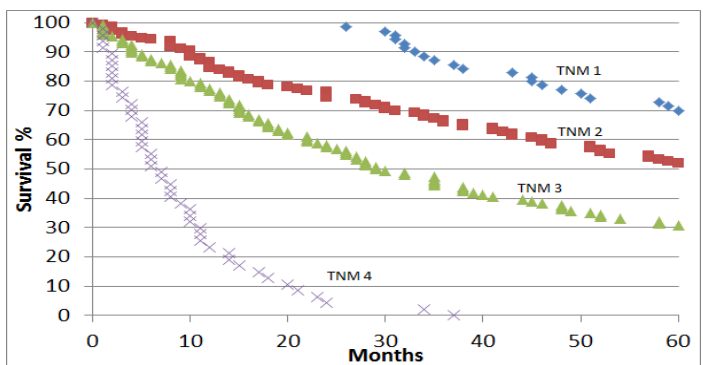
**Figure 4.** Survival Curves for patients at all 4 TNM stages

Again focusing on just TNM stage 2 and 3 patients we attempted to predict survival using both AI techniques and the TNM stage itself. The term "survival" is somewhat subjective but for the purposes of this work we used 5 years as the threshold for survival. Using the TNM stage alone to predict survival gives an accuracy of 64.6% (155 correct from 240). This is achieved by stating that all type 2 tumour patients will survive and all type 3 will not. If we use the cell marker statistics with patient physical details (such as age and sex) we can achieve a slightly higher predictive performance of 65.4% on the unseen test by using the top 8 attributes (table 1) selected by SVM attribute selection [12] followed by a trained Neural network [4]. As well as the physical characteristic of age, these 8 also included T Cell levels for CD24 and Tstroma and chemicals such as Interleukin 17 (IL17) signalling the expression of certain genes such as P27. The SVM based attribute selection method uses recursive feature elimination where the SVM is trained multiple times and the attributes subsequently evaluated post-training. We have already shown that this dataset shows anti-learning properties [8], whereby predicting TNM stage from immunohistochemical markers is best tackled using an anti-learning approach. An anti-learning approach [9, 10], using the bottom 6 attributes from the SVM attribute selection approach (table 1) achieves a 61.3% prediction rate. This included indicators such as the patients sex and levels of antibodies like clxcl10 and cxcr4.

These three approaches use different attributes to perform their prediction and there are many inconsistencies between the predictions. This can be used to our advantage by measuring performance, in an ensemble manner, on patients where 2 or all 3 of the models agree. This agreement will be with a subset of patients so will not apply to every patient, but the ability to give SOME patients a BETTER prediction of survival is extremely valuable. Conversely, where there isn't agreement the possibility of being less confident about a prediction could also be useful. Table 2 shows the individual performance of each model and also the performance when models agree. For example, when the simple TNM model and the neural network model for the top 8 attributes agree, we achieve a 79.2% prediction for survival. This can be viewed logically as:

**If Learned Neural model predicts a patient will survive 5 years AND Patient is TNM type 2**
   **OR**
**If Learned Neural model predicts a patient will not survive five years AND Patient is TNM type 3**
   **THEN**
**We can be ~80% confident that our prediction is correct**
   **ELSE**
**Patient survival cannot be predicted with any reliable accuracy**

The trained neural network used for this predictive task is shown in figure 5.

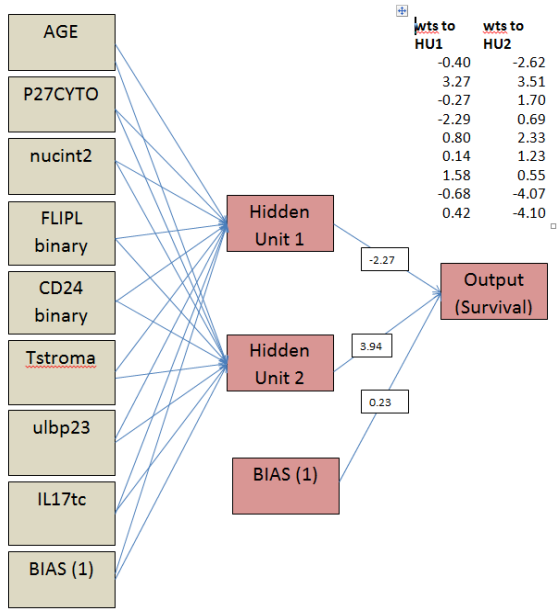

**Figure 5.** Structure and weights of a survival predicting artificial neural network.

While the non-linearity of a neural network decision making is notoriously difficult to glean information from, the relative simplicity of this neural approach means equation synthesis [13] yields formulas with some qualitative value, achieving predictive values of ~60% where models agree

| SVM Ranking | |
|---|---|
| (Top 8) | (Bottom 6) |
| nucint2 | betcytop |
| p27cyto | sex |
| age | b2m |
| IL17tc | trailint |
| FLIPL | cxcl10 |
| CD24 | CXCR4 |
| ulbp23 | |
| Tstroma (Peri tumour CD3) | |

**Table 1.** Top 8 and bottom 6 attributes as produced

| | Accuracy | Patients |
|---|---|---|
| TNM (T) | 64.6% | 240 |
| Learning (L) | 65.4% | 240 |
| Anti-learning (A) | 61.3% | 240 |
| T + L | 79.2% | 130 |
| T + A | 65.4% | 121 |
| L + A | 71.8% | 103 |
| T + L + A | 82.5% | 57 |

**Table 2.** Accuracy of models and subsets of patients by SVM attribute selection [12]

The dataset can now be divided into 4 distinct subgroup, each with a corresponding survival rate:

1. TNM 2 Patients predicted to survive by the learning model (%)
2. TNM 2 Patients predicted **not** to survive by the learning model (%)
3. TNM 3 Patients predicted to survive by the learning model (%)
4. TNM 3 Patients predicted **not** to survive by the learning model (%)

One interesting observation about this is that survival of TNM 3 patients with a positive prognosis from the model have a better survival rate that TNM 2 patients with a negative prognosis. This is shown more clearly by plotting the 4 corresponding survival curves (fig. 6)

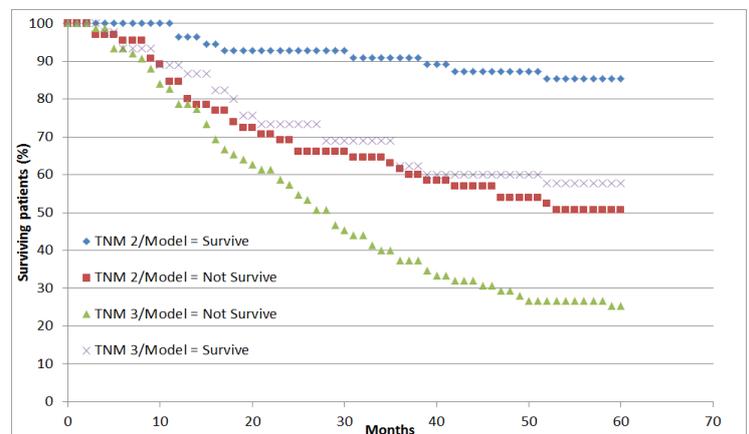

**Figure 6.** Survival curves for 4 distinct patient sets grouped on TNM stage and model prognosis

## V. CONCLUSIONS

We have presented results for a unique dataset based on the biochemical and factors associated with colorectal tumour patients. This dataset is limited in many ways, but extremely important nonetheless and modeling any relationships or features based on the dataset to hand is an urgent priority. Generally, whether attempting to predict TNM stages or survival, patients at TNM stage 1 and 4 have more clear indicators in the attribute set. TNM stage 2 and 3 provides a much more challenging prediction task, so much so that the TNM stage appears much less important when predicting survival for these 2 stages than other indicators.

The dataset used here has several drawbacks when it comes to building robust cause-effect models. The 2 main issues are the missing values and the high dimensionality. By pre-processing the dataset to convert attributes that may have a non-linear effect into linear attributes missing values can be represented as means of all values with a more stable meaning. This pre-processing step also makes resulting machine learning approach more capable of reaching global minima. The second issue of high dimensionality is tackled by using an attribute selection approach first demonstrated on another cancer dataset. We not only use the most highly ranked attributes here but by using anti-learning on the lowest ranked attributes we can produce a another, distinct model of the data.

When looking specifically at survival, using 3 unique predictive approaches allows us to compare predictions from all three approaches. The amount of differences between the 3 approaches could be inferred from the lack of relationships between TNM staging and the immunohistochemistry [8] and the lack of a relationship between unsupervised clusters and TNM stages [11]. So for the 240 patients, there was only agreement between 2 of the approaches for between 103 and 140 patients but when agreement occurred, predictive performance increased up to a maximum of 82.5%. This level of accuracy for predicting survival of patients with TNM stage 2 and 3 tumours is unprecedented, albeit on a small subset of patients. This approach offer an important opportunity for clinicians to improve prognosis for patients but also, any information gained by analysing the agreeing models can be fed back to researcher in their endeavours to improve treatment options at a cellular level.